\documentclass[pdflatex,sn-mathphys]{sn-jnl}

\jyear{2021}%

\theoremstyle{thmstyleone}%
\theoremstyle{thmstyletwo}%

\theoremstyle{thmstylethree}%

\usepackage{tabularx}
\usepackage{adjustbox}
\usepackage{amsmath}
\usepackage{parskip}
\usepackage{float}
\restylefloat{table}
\usepackage{xcolor}

\begin{document}

\title[Rivalry RL]{Incorporating Rivalry in Reinforcement Learning for a Competitive Game}


\author*[1]{\fnm{Pablo} \sur{Barros}}\email{pablo.alvesdebarros@iit.it}

\author[2]{\fnm{{\"O}zge Nilay} \sur{Yal{\c{c}}{\i}n}}\email{oyalcin@sfu.ca}

\author[1]{\fnm{Ana} \sur{Tanevska}}\email{ana.tanevska@iit.it}

\author[1]{\fnm{Alessandra} \sur{Sciutti}}\email{alessandra.sciutti@iit.it}

\affil*[1]{\orgdiv{CONTACT Unit}, \orgname{Italian Institute of Technology}, \orgaddress{\city{Genova}, \country{Italy}}}

\affil[2]{\orgdiv{School of Interactive Arts and Technology}, \orgname{Simon Fraser University}, \orgaddress{\city{Vancouver}, \country{Canada}}}


\abstract{Recent advances in reinforcement learning with social agents have allowed such models to achieve human-level performance on certain interaction tasks. However, most interactive scenarios do not have performance alone as an end-goal; instead, the social impact of these agents when interacting with humans is as important and largely unexplored. In this regard, this work proposes a novel reinforcement learning mechanism based on the social impact of rivalry behavior. Our proposed model aggregates objective and social perception mechanisms to derive a rivalry score that is used to modulate the learning of artificial agents. To investigate our proposed model, we design an interactive game scenario, using the Chef's Hat Card Game, and examine how the rivalry modulation changes the agent's playing style, and how this impacts the experience of human players on the game. Our results show that humans can detect 
specific social characteristics when playing against rival agents when compared to common agents, which affects directly the performance of the human players in subsequent games. We conclude our work by discussing how the different social and objective features that compose the artificial rivalry score contribute to our results.}


\maketitle

\section{Introduction}\label{sec1}

The social aspects of interaction are usually overlooked when optimizing an artificial agent through reinforcement learning \citep{liu2019deep}. Most of the training loop is done in an offline manner, 
or focuses on optimizing objective metrics that do not directly involve social aspects, for instance by using planners \citep{modares2015optimized} or human annotation feedback \citep{churamani2017teaching}. 
Some common success metrics in this regard involve 
solving the task in fewer steps, reducing predicted values, or achieving some predefined intermediate objective goals. When interaction with humans is the main goal, these artificial agents are evaluated mostly based on their objective performance \citep{modares2015optimized}. In the few examples where humans are present in the loop, the success measures are mostly related to the embodied interaction \citep{papaioannou2017combining, gao2019learning}, and not to the underlying decision-making process that these agents learned.

One scenario where these problems are very evident is 
competitive interaction. In a competitive game, 
an agent can learn to adapt towards its opponents by using reinforcement learning \citep{milani2020minerl}, even when these opponents are humans \citep{vinyals2019grandmaster}. However, it is extremely difficult to measure the social aspects of this interaction, without relying on typical human-robot or human-computer interaction schemes \citep{choudhury2019utility}. Although providing important insight on some social aspects, these evaluations usually focus on controlled lab-scenarios \citep{khamassi2018robot} and on the production of different robotic behaviors \citep{ritschel2017real} and dialogues \citep{cuayahuitl2019ensemble}. This in turn neglects exploring how the agents' various learning strategies influence their explicit behaviours and their interaction with humans \citep{tabrez2019improving}, despite it being one of their most important characteristics. This can be evidenced even in the new area of explainable reinforcement learning \citep{madumal2019explainable, sequeira2020interestingness}.

In this study, we address the problem of including social aspects in the learning strategies of artificial agents in a competitive scenario. We propose an objective human-centered metric based on rivalry \citep{havard2020rivalry}, to compose the reward function of the agents. Rivalry is a subjective social relationship arising between two actors, based on the competitive characteristics of an individual, as well as the increasing stakes and psychological involvement in the situation \citep{kilduff2010psychology}. We chose rivalry as it showcases the competing relation between individuals, which often affects their motivation and performance during gameplay \citep{kilduff2010psychology, kilduff2014driven}.  We model rivalry as a function of objective factors (such as game performance) and subjective information (such as certain personality traits and competitiveness level), and evaluate our model using the Chef's Hat Card Game \citep{barros2021s}


To obtain the social features of rivalry and map the intrinsic personality traits arising from human perception of learning agents, we run first an exploratory experiment where human players face artificial agents implemented using Deep Q-Learning (DQL) \citep{van2016deep} and Proximal Policy Optimization (PPO) \citep{schulman2017proximal}. Both learning agents are implementing COPPER \cite{barros2021you}, a continual learning adaptation for Chef's Hat agents. Using questionnaires, we collect how these agents impact the human players in terms of competitiveness, and how humans perceive the social characteristics of these agents. 


Using the compiled information from this experiment, we measure how humans perceive such agents in terms of rivalry. We then attach different social characteristics to each artificial opponent and use them in a rivalry synthesizing mechanism, to calculate the rivalry of an agent towards an opponent. We then run two ablation studies to develop a social characteristic predictor that will be used by the agents both when perceiving their opponents, and for finding the best manner to integrate rivalry in the learning routines.


Finally, we run a second experiment, where each artificial agent synthesizes a rivalry score against human players. By collecting the same information from questionnaires as in the first experiment, we can contrast the impact of the rival agents on the game when compared with non-rival agents.

Our results demonstrate that both learning agents are perceived as different social agents when playing against a human, in particular when compared with the random agent. When the learning is modulated by the rivalry score, we observe a strong contribution of rivalry on the performance of the human players. We discuss these results in terms of the contribution of the social and objective features for the formation of the rivalry function, and how they impact human perception. Ultimately, we detail how the performance of each agent changes when using the rivalry modulation.

\section{Related Work}

Reinforcement learning has received ample attention in the last years, in particular on the development of artificial agents. However, understanding the social role and the derived impact of social interaction within a learning mechanism is not yet fully explored. In particular, in multi-agent competitive scenarios, there is still a focus on performance-based metrics, which make it difficult to summarize, or even to verify, the social components which are directly affected when these models are deployed in scenarios involving humans. In this section, we detail the most relevant literature in this field, and the ones we based our entire explorations on.

\subsection{Reinforcement Learning in Competitive Games} In the late 1990s, several researchers tried to identify the impact of the Deep Blue artificial chess player \citep{campbell2002deep} on the development of artificial intelligence \citep{decoste1997future, decoste1998significance}. They all argue that beyond the technical challenge of beating a human, there is an underlying impact on how this agent affects the opponents' behavior during the entire interaction. Over time, these investigations were let aside by the mainstream community, which focused mostly on solving more complex problems. This vision is reflected by the recent development of deep reinforcement learning and the research on training artificial agents to play competitive games that have flourished since \citep{shao2019survey}. AlphaGo \citep{silver2016mastering} demonstrated that these agents can play competitively against humans in very complex games. The recent development of agents that play the StarCraft computer game \citep{vinyals2019grandmaster} pushes these boundaries even further. These agents learn how to adapt to dynamic environments, how to map hypercomplex states and actions, and how to learn new strategies \citep{shao2017cooperative}. Most of the studies however focus on the final goal of these agents: how to be competitive against humans. As such, none of them focus on understanding the impact that these agents have on human opponents.


\subsection{DQL and PPO Playing Behavior on the Chef's Hat Card Game} In the same development wave, it was recently investigated the design and development of reinforcement learning agents to play the four players Chef's Hat competitive card game \citep{barros2020ad}. These agents were based on Deep Q-Learning (DQL) \citep{van2016deep} and Proximal Policy Optimization (PPO) \citep{schulman2017proximal}, and achieved success in learning how to win the game in different tasks: playing against random agents, self-play, and online adaptation towards the opponents. However, it was observed that these agents present different behavior during gameplay, while maintaining a similar objective performance measured by overall wins over a series of games. What has not yet been done is to measure the social impact that such strategies can have when the agents play against humans.


\subsection{Human-centric Analysis of RL} When analyzing the impact of artificial agents on humans, there are now decades of studies focused on Human-Robot \citep{rossi2017user} and Human-Computer \citep{arzate2020survey} Interaction (HRI and HCI respectively). Social and affective computing research suggests humans tend to treat computers as social actors, \citep{reeves1996media} where attributes such as personality and emotions are modeled to affect how these agents are perceived \citep{ball2000emotion, zhang2016modeling}. In an interaction setting, these attributes can be used to change the behavior of humans or improve interaction in different contexts \citep{leite2013social, moerland2018emotion}.
In RL, most of these studies however focus on optimizing RL agents to solve a specific task, even social ones, without having much feedback on the social aspects of the task as part of their learning mechanism. Such agents are usually designed to learn an expected outcome, such as improving engagement \citep{papaioannou2017combining}, or imitating humans \citep{milani2020minerl}. None of the most recent studies focus on extracting the intrinsic behavior bias that different learning schemes apply to the final agent.

\subsection{Rivalry in Cooperative and Competitive Games} One way to explain the behavior of an agent as a factor of its learning strategy is to measure its impact on humans. In cooperative scenarios, there exist several social metrics that take interaction into consideration \citep{murphy2013survey, perolat2017multi}, but most of them focus on the subjective impressions humans have of the embodied interaction \citep{canamero2020embodied}, the subjective quality of the interaction \citep{peltason2012talking}, or the efficiency of the interaction to solve a task \citep{begum2015measuring}. 

In a competitive game however, one of the most informative metrics is the rivalry \citep{havard2020rivalry} between the human and the agents. Rivalry is defined as a competitive relation between individuals or groups, characterized by the subjective importance placed upon competitive outcomes (i.e., win or lose) independent of the objective characteristics of the situation (e.g., tangible stakes) \citep{kilduff2010psychology, kilduff2014driven}. A proposed theoretical model of rivalry suggests that antecedents of rivalry are similarity factors, competitiveness, and relative performance of the agents \citep{kilduff2010psychology}. The presence of a rivalry effect, in turn, affects the motivation of the individual and their performance \cite{kilduff2014driven}. We aim to evaluate how different agents affect the user's perception 
and their performance due to the increased competitiveness and rivalry effects. 

In competitive games, rivalry is a central concept that directly affects the opponent's behavior through their motivation in play. In human-to-human scenarios and economics, a healthy rivalry is considered to be an important factor that can positively affect the performance of opponents; while in other situations, it can also contribute to unnecessary risk-taking behavior \citep{kilduff2010psychologythesis}. 

In human-in-the-loop online learning scenarios for competitive games, the absence of rivalry or competitiveness might result in the human opponent to lose motivation to play the game and show sub-optimal performance during gameplay. The agents who are learning actively from the human, are bound to learn from this bad performance input, which in turn would result in sub-optimal learning. 
By introducing the notion of rivalry, 
we expect to increase motivation and engagement in the game and thus also the user performances.

\section{Proposing Artificial Rivalry}

Our rivalry modulation acts directly on two types of agents: A Deep Q-Learning (DQL) one and a Proximal Policy Optimization (PPO) one. Both agents were recently adapted and optimized for the Chef's Hat game through the COPPER modulation \cite{barros2021you}. COPPER introduces an opponent-specific experience-prioritizing memory used to improve the continual learning capabilities of each agent when playing against known opponents. We use these agents' implementation as our artificial opponents, and apply the rivalry modulation to their learning mechanism. It is important that our agents continually update their playing strategy when playing against the human opponents, so the rivalry score is created and updated accordingly.


\subsection{A Chef's Hat Agent}

The DQL and PPO implementations of the agents were chosen due to their success on learning different strategies \cite{barros2020ad}, and their good performance when playing against human players \cite{barros2021you}. Both agents are implemented as COPPER-based agents, and are set to keep learning during all of our experiments. 

\begin{figure}[htbp]
    \centering
    \includegraphics[width=0.8\columnwidth]{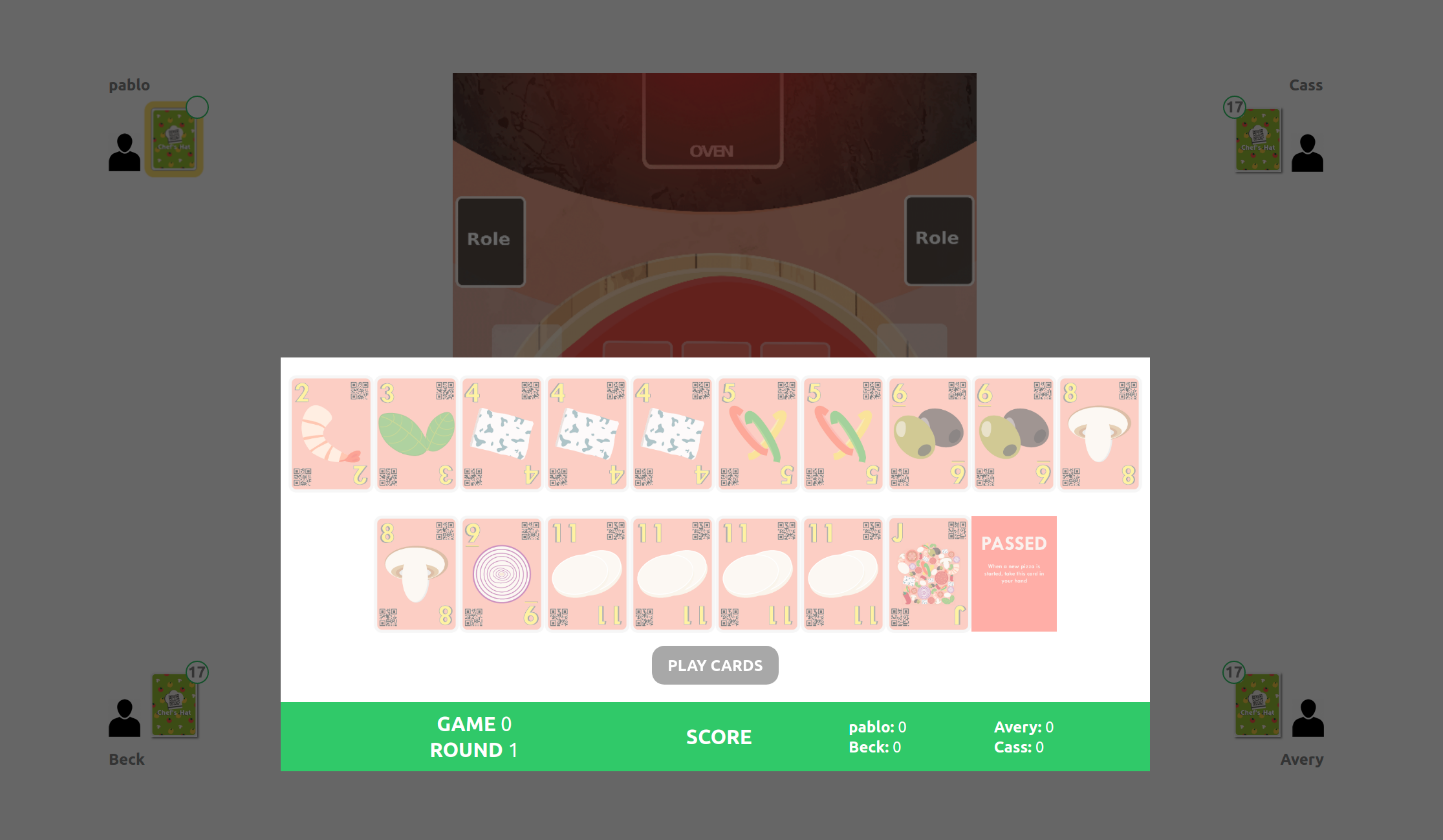}
    \caption{Illustration of the Chef`s Hat card game environment used for all our experiments. Available at: https://github.com/pablovin/ChefsHatGYM}
    \label{fig:chefsHat}
\end{figure}

The Chef's Hat game, illustrated in Figure \ref{fig:chefsHat}, is a multiplayer competitive card game. At the beginning of the game, each player receives 17 cards, and the player that discards all of them first wins the match. The entire rules of the game and the capability of agents to learn different strategies were recently explored in different studies  \cite{barros2020ad, barros2021s, barros2021you}. The game state is composed of 28 values, referring
to the 17 possible cards each player has in their hands, and 11 cards on the game board. The action space is represented by 200 different discard actions an agent can do, which reflects the complexity of the strategy formation in this game.

\subsubsection{Deep Q-Learning}

Our Deep Q-Learning (DQL) agent implements a $Q$ function as:

\begin{equation}
Q:SxA \rightarrow \mathbb{R}
\end{equation}

where $S$ is the state, in our case represented by the 28 values composed by the cards at hand and the cards at the board. The actions, $A$, are expressed using the 200 discrete values for all the possible actions.

To update the Q-values, the algorithm uses the following function:

\begin{equation}
\label{Eq1}
    Q'(s_t, a_t)= Q(s,a) + \alpha  \times \left (  TD \right )
\end{equation}

where $t$ is the current step, $\alpha$ is a pre-defined learning rate and $TD$ is the temporal difference function, calculated as:

\begin{equation}
\label{Eq2}
TD =r_t \times \gamma \times maxQ \left ( s_{t+1, a_t}  \right ) - maxQ \left (s_{t1, a_t}  \right )
\end{equation}

where $r_t$ is the obtained reward for the state ($s_t$) and action ($a_t$) association, $\gamma$ represents the discount factor, a modulator that estimates the importance of the future rewards, and $maxQ\left (s_{t1, a_t}\right )$ is the estimate of Q-value for the next state.

This agent also implements a target model, which is a time-delayed policy which receives a snapshot of the original policy after a certain number of training steps.

The target model is used to obtain the target Q learning when calculating the updated $TD$:

\begin{equation}
\label{Eq3}
TD =r_t \times \gamma \times maxQ\left (s_{t+1, a_t}  \right ) - maxQ_t\left (s_{t1, a_t}  \right )
\end{equation}

where $maxQ_t\left (s_{t1, a_t}  \right )$ is the Q-values obtained from the target network.

\subsubsection{PPO}

Our Proximal Policy Optimization (PPO) \cite{schulman2017proximal} agent implements an actor-critic model:

\begin{equation}
\label{Eq4}
A(s_t,a_t) = r_t + \gamma \times V(s') - V(s)
\end{equation}

where $V(s)$ represents the critic value for a given state, and $V(s')$ for the future state. The advantage function is used to stabilize training the actor-network, while the critic network uses the discounted rewards as the target. 

The actor-critic base model is updated by implementing adaptive penalty control, based on the Kullback–Leibler divergence, to drive the updates of the agent at each interaction. It showed to be important to learn different and more optimized strategies \cite{barros2020ad}, probably due to how this method simplifies the necessity of having a large-memory replay \cite{bansal2017emergent, kidzinski2018learning}.

\subsection{COPPER}

Both agents implement the COPPER modulator \cite{barros2021you} that expands the prioritize experience replay ($PER$). Traditional $PER$ can be expressed as:

$PER(i)$ is calculated based on the network's loss after calculating $TD$ in a forward pass of the network (using an input $i$):

\begin{equation}
\label{Eq5}
PER(i) = \frac{p_{a}^{i}}{\sum_{k}p_{a}^{k}}
\end{equation}

where $a$ indicated how much we want to rely on the priority, $p$ is the priority, and $k$ the total number of saved experiences. COPPER, on the other hand, introduces a new opponent-specific term ($o$):

\begin{equation}
\label{Eq6}
COPPER(i) = o\frac{p_{a}^{i}}{\sum_{k}p_{a}^{k}}
\end{equation}

In recent experiments, COPPER was shown to be much more effective on learning new strategies against recurrent opponents, in particular when playing against humans \cite{barros2021you}.

\subsection{Modeling Rivalry From a Human Perception} \label{section3}

\begin{figure}[htbp]
\centering
\includegraphics[width=0.5\columnwidth]{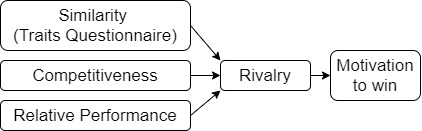}
\caption{The theoretical model of rivalry used for our new rivalry metric proposition, based on the framework proposed by \citep{kilduff2010psychology}.}
\label{fig:rivalry}
\end{figure}

To optimally define the impact each agent has on the players, we used a standard formalization of Rivalry \citep{kilduff2010psychology}. Rivalry can be defined as a subjective social relationship arising between two actors based on the competitive characteristics of an individual, as well as the increasing stakes and psychological involvement in the situation. Thus, a proposed theoretical model of rivalry, illustrated in Figure \ref{fig:rivalry}, suggests that antecedents of rivalry are similarity factors ($S_a$), competitiveness ($C$), and relative performance ($P$) of the agents \citep{kilduff2010psychology}.

Rivalry research suggests that individuals tend to evaluate their abilities by comparing their performance to persons who have similar characteristics to themselves. The similarity can be measured in demographics \citep{kilduff2010psychology}, gender \citep{miller1984self}, personality \citep{kilduff2010psychology}, perceived traits \citep{tauer1999winning} and rank in competition \citep{kilduff2014driven} to assign social or behavioral attributes to the other.

For this work, in addition to the ranking and performance similarity, we included trait similarity as a factor 
consisting of competence, agency, and communion traits as the indicators of relevant behavioral attributes. These traits are frequently used in social stereotypes research (e.g., gender, nationality, age, status) \citep{Eagly_Nater_Miller_Kaufmann_Sczesny_2020} and have been shown to affect liking \citep{wojciszke2009two}. Moreover, social perception \citep{cuddy2008warmth} has been previously shown to affect competitive behavior \citep{fiske2002model}. Agency, competence, and communion traits have also been used in virtual agent research to examine users' judgments of expected agent behavior \citep{nag2020gender}. These traits were chosen as representative ones due to their comprehensibility, which facilitates the self-and other assessment \citep{abele2007agency}. We collect the agency ($ag$), competence ($ct$), communion ($cm$), and competitiveness ($C$) assessments of each of the agent players, from a human perspective, through an exploratory experiment. Once these values are calculated, we can define a rivalry score as ($R_a$):

\begin{equation}
\label{Eq7}
S_a =  \sqrt{(ag_h-ag_a)^{2} + (ct_h-ct_a)^{2} + (cm_h-cm_a)^{2}}
\end{equation}

\begin{equation}
\label{Eq8}
C = C_h 
\end{equation}

\begin{equation}
\label{Eq9}
P_a =  (points_h - points_a) / 15
\end{equation}

\begin{equation}
\label{Eq10}
R_a =  (S_a + C_h + P_a) / 3
\end{equation}

\noindent where the index $a$ defines one of the opponents, the index $h$ defines the human ratings about the social behavior of the opponents, obtained in a prior human study. The individual performance of each player is calculated by the Chef's Hat environment \footnote{https://github.com/pablovin/ChefsHatGYM}
as the sum of the average score of each game (given by the sum of the points in each round, divided by the number of rounds in the game), divided by the total number of played games.

The rivalry score ($R_a$) is used as our main evaluation metric on how the agents impact the human players, thus we expect it to change accordingly to the game development. As the participants evaluate the entire game behavior of an agent, and the only difference among artificial players is the way they play the game, the measure of rivalry reflects directly the user's perception of the outcome of the reinforcement learning strategies.

\subsection{Rivalry as a Learning Modulator}

Once rivalry can be defined from a human perspective, we need to model it from an agent perspective, and use it as a learning modulator. This will make the agents create a rivalry sense against the other players, and most important, will help the agents to develop a behavior that can make it be identified by a human as a rival.

To achieve a rivalry modulator, we calculate rivalry in a similar manner as we do from a human perspective. To predict similarity, though, the agents used a similarity predictor trained based on the humans' responses collected during our exploratory experiment. The agent-player similarity predictor ($pr_a(h)$) matches state+action chosen by humans ($h$) with the human-assessed competence, agency and communion traits ($ag_a$, $ct_a$ and $cm_a$), and it follows the survey scales used in virtual agent research in terms of calculation of the scores for each trait \citep{nag2020gender}. 

Our similarity predictor was built as a Multi-Layer Perceptron (MLP) neural network mapping an interval of action/spaces with a set of similarity scores. The similarity predictor of each agent inferred during gameplay the traits of each human the agents played against. Each type of agent (DQL and PPO) also have a single set of traits associated, derived from the human judgments provided in the exploratory study. Thus, the similarity based on an agent's ($a$) perspective ($S_h$) is given as:

\begin{equation}
\label{Eq11}
S_h = \left ( \sqrt{(pr_a(h) - (ag_a, ct_a, cm_a) )^{2}} \right )
\end{equation}

The performance measure of the agents was given by their own assessment of their actions. To achieve this, each agent computed the introspective confidence ($ic_a$) \cite{cruz2020explainable} of each action, which focused on scaling the selected Q-value of an action towards the final goal, using a logarithm transformation which computes the probability of success. In our case, this was the probability of winning the game. The introspective confidence gives us a self-assessment of the agent's actions, based on its own game experience. We use the introspective confidence as the agent's competitiveness:

\begin{equation}
\label{Eq12}
C_a = \frac{\sum ic_a (act) }{totalActions}
\end{equation}

\noindent where $act$, is each of the actions the agent took during the game.

The relative performance ($P_h$) is calculated similarly to the human's perspective but considers the agent's perspective. Thus, the predicted rivalry ($R_h$) is defined as the mean of  the previous three factors:

\begin{equation}
\label{Eq13}
R_h = (S_h + C_a + P_h).
\end{equation}

To guarantee the agent learns how to be a rival of an opponent, we include the predicted rivalry ($R_h$) into the final reward of the agent by a simple weighted average, optimized in an ablation study described below.

\section{Experimental Setup}

Our rivalry modulation is directly related to a very specific reinforcement learning task: multi-agent competitive interaction. In this regard, the Chef's Hat Card game was chosen as our primary investigation environment. As our experiments involve a good mix of human-based studies and artificial agents optimization, we separate them into four categories: first, we perform our exploratory study to understand how the PPO and DQL agents are perceived by humans. Our second experiment uses the information collected from the exploratory study to train and validate the similarity predictor neural network. The third experiment optimizes and evaluates the rivalry learning modulator. And finally, we run a human-based study to identify the impact of the rival agent on human perception.


\subsection{Chef's Hat Environment}

The Chef's Hat Environment is an OpenAI GYM-based implementation of the Chef's Hat card game \footnote{https://github.com/pablovin/ChefsHatGYM}. It includes all the rules and mechanics of the original game and can be used to train and evaluate artificial agents. The game itself is based on turns, and on each turn a player can do a discard action or a pass action. For every match played, the players gain points based on their finishing position, with the winner gaining 3 points. A full game consists of several matches until one of the players reaches 9 points.

Following our previous experiments with the Chef`s Hat environment, we use a global reward scheme that only gives a full reward once the agent performs an action that leads to it winning the game. We also start the game with pre-trained agents, which learn how to play the game using a self-play strategy \cite{barros2020ad} available through the Chef's Hat Player's Club repository \footnote{https://github.com/pablovin/ChefsHatPlayersClub}.

To collect human data, and to allow humans to play against the agents, we use the Chef's Hat Online software \footnote{https://github.com/pablovin/ChefsHatOnline}, which is a browser-based interface for the Chef's Hat Environment and allows human players to participate in the game.

\subsection{Exploratory Study: Understanding the Agents }

In our first experiment, we perform an exploratory study using the Chef`s Hat Online environment. The goal of this study is two-fold: we want to understand and measure the impact of each learning agent on the rivalry attribution. Also, we collect the human attributions to these agents, and use them to describe both agents socially, when synthesizing rivalry.

For this study, we implement three agents, a DQL, a PPO, both with COPPER, and a naive agent that only performs random actions during the entire game. A human plays a 9-points game against these agents, and we collect the entire game status over the entire experiment, which includes all players' performance. We also run two questionnaires, one at the beginning of the game to collect the human players' self-assessment, and one by the end of the game to investigate the participants' perspective about the agents.

In our questionnaire, we follow the standard traits questions to measure competence, agency, and communion \citep{Eagly_Nater_Miller_Kaufmann_Sczesny_2020}. In particular, we measure competence score as an average of the scores of the items related to ambition, courage, decisiveness, and aggressiveness; agency as an average of intelligence, innovation, organization, and compassion; and communion as an average of compassion, affection, emotional response, and sensitiveness. Each of these terms is represented with a Likert scale that varies from 1 (Not at all) to 5 (Very). 

We also asked humans to self-assess their competitiveness and the perceived competitiveness of each agent, once the game was over. Each agent was only identifiable by one of three names (Evan, Dylan and Frankie), in order to make the entire evaluation based  on their game behaviour alone. The questionnaires are available on our Appendix. In total, 28 different persons played the game.

Using the collected information, we calculate the rivalry score from the human perspective for each agent. We then proceed with a statistical test to identify the contribution of each of the terms that compose rivalry (similarity,  relative performance and competitiveness).

\subsection{Rivalry Ablation: Similarity Predictor}

The goal of this experiment is to achieve a reliable similarity predictor, to be used by the agents. We implement it as a fully-connected multi-layer perceptron (MLP) neural network, that receives as input a sequence of pairs of action and cards on board, and predicts a similarity label. We fine-tune it and evaluate it using the data obtained from the previous  exploratory study.

To optimize this network we used a Tree-Parzen Optimizer (Hyperopt \citep{bergstra2013hyperopt}), and a cross-validation with 70\% of the collected data for training and 30\% for testing. We report the optimization search-space and the final architecture in the Appendix Section A.3. We calculate the performance in terms of mean accuracy over 30 runs.

\subsection{Rivalry Ablation: Rivalry Learning Modulator}

To maximize the effect of the rivalry modulator in the agent's learning, but without losing the focus on winning the game, we run an optimization study to find the best weight when adding rivalry in the agent's reward function. Our rivalry score is defined in a way that it would increase when an agent plays against itself, as the social behavior, performance, and game strategy of the agents are the same. So, we optimize the rival weight towards the reward by maximizing the rival score, while maintaining a similar performance against each other.

We run 1000 simulation games, where an agent plays against three other agents, among which one is another instance of itself. During all the games, we add the rival score weight as an updatable parameter of the network, and optimize it towards the opponent that has the same implementation of the agent. We stop the training when the rival score of both agents' instances towards each other is maximized. We track the rivalry evolution over time, together with the agents' performance, to guarantee that the optimization reaches a satisfactory state.

\subsection{Rivalry Impact: Playing Against Humans}

Once the rival agents are implemented, and we found the optimal reward weight, we deploy them into the same scenario as our first exploratory experiment. We repeat the same settings, but now the human play the game against a rival agent, a non-rival COPPER agent that continues learning and a non-rival agent that does not learn during the game. In this experiment, as we are interested on the differences within these agents, we only implemented a DQL-based agent. The goal of this experiment is to verify the impact of the rivalry in terms of perceived game play. We collect the same data, using the same questionnaires, and calculate the rivalry score of all the participants towards the agents.

We then run statistical tests to identify the impact of the rival agent in the game, in particular when compared to the non-rival agents. This experiment aims to identify the contribution of rivalry in terms of social perception, but also on the final performance of the agents.

\section{Results}

\subsection{Exploratory Study}
A total of 28 games were completed and were therefore included in all subsequent evaluations in this experiment. Of these 
games 
13 were played in English, 14 in Portuguese, and 1 in Italian. All of them, however, played using the same Chef's Hat Online platform, so following the same game rules. Of the participants who played these games, 54\% were between the ages 31-50, 36\% reported to be between 18-30, 7\% reported to be aged over 50, and one participant chose not to disclose their age.

On average, each game was played for 3.34 matches (SD=.47), with an average score of 2.25 (SD=.71). We calculated the rivalry scores for each participant towards each of the three agents, based on the proposed equations (see Section \ref{section3}), by using the Similarity, Competitiveness and Performance measures.

Similarity scores were computed using the Equation \ref{Eq8} (see Section \ref{section3}). 
An analysis of the results showed that agents were rated as significantly different in terms of their overall similarity scores with participants (Friedman test: ${\chi}^2$=6.82, df=2, p=.033). Further analysis using Durbin multiple comparisons \citep{pohlert2018package} showed that the random agent was perceived significantly different than the DQL (p=.009) agent. Moreover, there were significant differences in the similarity scores for particular items in the traits questionnaire, namely ``Decisive" (Friedman test:${\chi}^2$=6.40, df=2, p=0.041) and ``Innovative" (${\chi}^2$=6.05, df=2, p=0.049) traits, and scores showed a tendency to differ for ``Creativity" (${\chi}^2$=5.43, df=2, p=0.066).

The relative performance scores stand of the three agents showed significant difference (${\chi}^2$= 28.1, df=2, p$<$ .001), where both the DQL (Durbin multiple comparisons, p$<$ .001) and PPO (p$<$ .001) agents were significantly different from the random agent. No significant difference was seen between the DQL and PPO agent (p=.23). Finally, we used the self attributed values for the competitiveness scores for all the users (M=.89, SD=.17). Rivalry scores calculated using these three measures were significantly different among the agents (${\chi}^2$=25, df=2, p$<$ .001). Pairwise comparisons revealed that both DQL (p$<$ .001) and PPO (p$<$.001) agents were significantly different than the random agent. No difference was found between the DQL and PPO agents (p=0.791). 

\subsection{Similarity Predictor}

In our First Experiment, we collected a total of 16,000 action/cards on board pairs, and associated them with the ``Decisive", ``Innovative" and ``Creative" labels self-assessed (from humans), or given by (from the agents) during the human data collection, using the average of them as our final similarity score. We report the optimized search-space and the final architecture in Table \ref{tab:searchSpaceSimilarityFinal}.

\begin{table}[H]
\center
\begin{tabular}{ |c|c|} 
\hline
 \textbf{Parameter} & \textbf{Search Space} \\ \hline
 Concatenated Actions & [3,\textbf{5},10, 15]\\
 Number of Layers & [1,\textbf{2},3]\\
 Units Per Layer & [16, 32, \textbf{64}, 128, 256, 512, 1024] \\
 \hline
\end{tabular}
\caption{Search space and final architecture, in bold, used to optimize the similarity predictor. }
\label{tab:searchSpaceSimilarityFinal}
\end{table}

The final performance of the similarity predictor after running 30 cross-validation routines, when trained with the data collected from humans and agents, is 83\% accuracy, with a standard deviation of 0.5.

\subsection{Rivalry Learning Modulator}

After running 1000 games, the final optimized weighted value for the reward is as follows: 

\begin{equation}
R = R_o + 0.2*R_h
\end{equation}

where $R_o$ is the original reward of the agent, and $R_h$ is the rivalry score. We also tracked the performance of the agent over time. Without rivalry, when playing in the same setting, a DQL agent obtained an average of 1.3 (game score), with a standard deviation of 0.1. When the rivalry modulation was used, the average wins stayed at 1.2, with a standard deviation of 0.2.

\subsection{Rivalry Impact}

In this experiment, a total of 25 games were completed. From the final set of the completed games, 9 were played in English, 9 in Portuguese, 3 in Spanish and 4 in Italian. From the participants who played these games, 55\% reported to be aged between 31-50 and the remainder reported to be between 18-30.

On average, the games were played for 3.65 matches (SD=.48), and with an average human score of 3.06 (SD=.68) which is significantly higher than the participants' scores from Study 1 (t(46)=-3.98, p$<$0.001). Similar to the previous study, the rivalry scores for each participant were calculated using the proposed equations in Section 3, by using the Similarity, Competitiveness and Performance measures.

An analysis of the similarity scores, computed using the Equation \ref{Eq8} (see Section \ref{section3}), showed that the agents were not rated as significantly different in terms of their overall similarity with participants (Friedman test: ${\chi}^2$=1.57, df=2, p=0.457). The relative performance scores of the three agents showed significant differences (${\chi}^2$= 23.7, df=2, p $<$ .001), where Durbin-Conover multiple comparisons test revealed that all agents performed significantly differently than each other. Pairwise comparisons showed the rival DQL agent performed the best, being significantly better than both COPPER DQL (p= .036) and offline DQL agents (p$<$ .001). The COPPER DQL agent was also significantly better (p$<$.001) than the DQL agent.

\begin{table}[H]
\center
\begin{tabular}{l |c|c } 

 Agent type & Performance (SD) & Rivalry (SD)  \\ 
 \hline
 Rival DQL  & 1.50 (.50) & 0.44 (.08) \\  
 COPPER DQL & 0.63 (.24) &  0.35 (.08) \\
 Offline DQL & 0.41 (.32) & 0.30 (.36)
\end{tabular}
\caption{Relative performance and rivalry scores of the three agents.}
\label{tab:agentPerformanceScores}
\end{table}

Finally, the rivalry scores calculated using these two measures and the self-attributed competitiveness scores of the participants (M=.79, SD=.18) were significantly different among the agents (${\chi}^2$=15.7, df=2, p$<$ .001). Pairwise comparisons revealed that the rival DQL agent had significantly higher rivalry scores compared to both the COPPER DQL (p= .036) and DQL (p$<$.001) agents. Further, the COPPER DQL agent had significantly higher rivalry scores (p=.009) than the DQL agent. However, the participants' motivation to play with agents were not significantly different among different types of agents (${\chi}^2$=0.2133, df=2, p=0.9). Table \ref{tab:agentPerformanceScores} shows the values for the relative performance and rivalry scores of the three agents.

\section{Discussions}

In our study, we are mostly interested on how humans perceive the impact of different learning strategies when interacting with artificial agents. In particular, we evaluate if we can modulate this perception in a controlled manner using the rivalry term. Our experiments demonstrate that we can indeed achieve such manipulation, although in a limited manner. In this section, we discuss these findings in more detail.

\subsection{Are Learning Agents Perceived as Rivals?}
Our experimental results provided insights to identify whether agents trained with different RL strategies  yield  distinct  rivalry and whether including rivalry in the reward function enable agents to modulate human's response on a rivalry scale. Our first exploratory study showed that agents trained with DQL and PPO strategies yield distinct rivalry if compared to a random agent when playing the Chef’s Hat card game against humans, which confirms that their capability of learning a strategy is indeed perceived by humans. However, we failed to find a significant difference between DQL and PPO agents, which mirrors the lack of significant difference between their relative performances and their scores for the ``Decisive", ``Innovative" and ``Creative" traits. 

\subsection{Agents Optimization}
\begin{figure}[htbp]
\centering
\includegraphics[width=1\columnwidth]{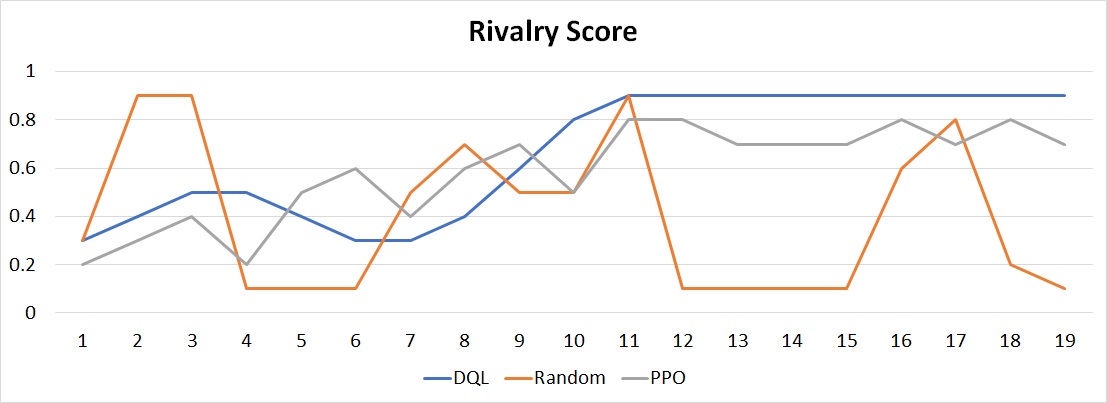}
\caption{Example of the rivalry score calculation of a rival DQL agent playing against a non-rival DQL, a PPO and a random-based agent.}
\label{fig:rivalryExample}
\end{figure}

When developing the agents, we needed to adapt them towards using rivalry as a learning modulator. Our experiments demonstrate that in our simulation environment, the inclusion of the rivalry term does not affect the general performance of the agent, and helps on changing their underlying behavior. This behavior can be explicitly demonstrated when a DQL rival agent plays a game against a non-rival version of itself, a PPO agent and a random agent. Figure \ref{fig:rivalryExample}, illustrates a match between a DQL agent, with rivalry, a DQL agent without rivalry, a random agent, and a PPO agent. We observe that the rivalry value increases constantly for the DQL agent, as they share the same strategy and social traits. Against the PPO agent, the rivalry score increases at a different pace, which demonstrates that the rival agent can detect the strategy of the PPO agent, and associate a different social trait set to it. A random agent however does not have a strategy, and the rival score fluctuates for each action it takes as expected.

\subsection{The Role of Rivalry in Chef's Hat}

Our last experiment, where we measure the impact of rivalry, showed that agents trained with a predicted rivalry as a part of their reward function can modulate human responses on a rivalry scale, and in turn yield significantly higher rivalry results.  Moreover, the rivalry scores for the rival, COPPER-based, and offline learning agents were all significantly different from each other, with the rival agent exhibiting the highest score. Similarly to the exploratory study, these differences mirrored the relative performances of the agents. The similarity results instead were not significantly different for any of the agents. This suggests that agents' behaviors were not providing enough information to be perceived as different in terms of their agency, communion and competence traits. 

Our experiments confirmed the importance of agent's performances to trigger a sense of rivalry. Moreover, the agents learned to modulate their behavior enough to yield a significantly higher rivalry.  However, their performances did not reach those of the participants, suggesting there is still room for improvement.
  
The addition of rivalry and the better performances of the agents seemed to increase the motivation of participants to play better, as it can be seen in the increased average scores from the first exploratory to the second one. Although we cannot exclude that this might depend also on a difference between the two players' samples, such result would be expected from rivalry literature \citep{kilduff2010psychology}. We saw that the significant difference in rivalry scores did not had a significant effect on participants' motivation to play with different agents. Motivation to play again could be further investigated by allowing participants to only play with one type of agent in future studies.
Another observation is that the rival RL agents were not associated to different judgments in terms of social traits such as agency, competence or sense of communion, on which the similarity estimation is based. This might be due to the scenario of the game, where the agent's behavior could be appreciated only from its choices in the game.
This point could be further investigated in a study where agents have more indicators of social traits such as an embodiment, gestures or emotional expressions.

However, even in such a socially impoverished setting, the rival agent succeeded in triggering a higher rivalry in the human participants, revealing the potentiality of introducing rivalry in social reinforcement learning research.

In comparison with the first human experiment, the subjects of the second human experiment played slightly more turns, but achieved a significantly higher average score. This might be an indication that the rival agent made the game more motivating, and more challenging. Observing these characteristics from the behavior of each agent alone is a strong indication that when adding explicit social traits to the agent's behavior, the rivalry modulation could be a game-changer on social reinforcement learning research.



\section{Conclusion}

In this work, we examined the inclusion of social aspects in the learning strategies of reinforcement learning (RL) agents in a competitive card game scenario. We proposed the social metric of rivalry, based on the background research in social psychology, and trained RL agents with a reward function that reflects this metric. The resulting agents, trained with the rivalry metric, were successful in yielding significantly different play styles, distinguishable in terms of a main antecedent of rivalry: relative performance. However, the similarity scores calculated based on social traits - another main factor in rivalry - failed to show differences. We plan to further address this issue in future studies by equipping RL agents with distinguishable social signals. Our results suggest that using social concepts such as rivalry shows promise in training agents to perform in human interaction contexts.

\section{Declarations}

The authors have no conflicts of interest to declare that are relevant to the content of this article.

\begin{appendices}

\section{Questionnaires}\label{secA1}

We used two questionnaires in our experiments. One before the game start, to collect the self-assessment information of the players, and one after the game finishes, to identify the players' perception of the agents. The trait similarity questions consisting of competence (Questions 9 to 12), agency (Questions 5 to 8), and communion (Questions 13 to 16) traits as the indicators of relevant behavioral attributes, as used as indicators of stereotypical traits \cite{Eagly_Nater_Miller_Kaufmann_Sczesny_2020} and previously used in virtual agent research \cite{nag2020gender}.

\subsection{Pre-game questionnaire}

\begin{enumerate}
    \item Which was the nickname you choose?
    
    \item What is your mother tongue?
    
    \item How old are you?
    \begin{itemize}
        \item Less than 18 years old
        \item Between 18 and 30
        \item Between 31 and 50
        \item More than 50
        \item Do not want to disclose
    \end{itemize}
    
     \item How competitive you are when playing games?
     
     \begin{table}[H]
         \centering
         \begin{adjustbox}{width=0.6\textwidth}
             \small
             \begin{tabular}{|c|c|c|c|c|}
               \hline    1(Not at all) & 2 & 3 & 4 & 5 (Very) \\ \hline
             \end{tabular}
         \end{adjustbox}
     \end{table}

    \item How ambitious are you when playing games?
     
      \begin{table}[H]
         \centering
         \begin{adjustbox}{width=0.6\textwidth}
             \small
             \begin{tabular}{|c|c|c|c|c|}
               \hline    1(Not at all) & 2 & 3 & 4 & 5 (Very) \\ \hline
             \end{tabular}
         \end{adjustbox}
     \end{table}

    \item How courageous are you when playing games?
     
      \begin{table}[H]
         \centering
         \begin{adjustbox}{width=0.6\textwidth}
             \small
             \begin{tabular}{|c|c|c|c|c|}
               \hline    1(Not at all) & 2 & 3 & 4 & 5 (Very) \\ \hline
             \end{tabular}
         \end{adjustbox}
     \end{table}

    \item How decisive are you when playing games?
     
     \begin{table}[H]
         \centering
         \begin{adjustbox}{width=0.6\textwidth}
             \small
             \begin{tabular}{|c|c|c|c|c|}
               \hline    1(Not at all) & 2 & 3 & 4 & 5 (Very) \\ \hline
             \end{tabular}
         \end{adjustbox}
     \end{table}

    \item How aggressive are you when playing games?
     
      \begin{table}[H]
         \centering
         \begin{adjustbox}{width=0.6\textwidth}
             \small
             \begin{tabular}{|c|c|c|c|c|}
               \hline    1(Not at all) & 2 & 3 & 4 & 5 (Very) \\ \hline
             \end{tabular}
         \end{adjustbox}
     \end{table}

    \item How creative are you when playing games?
     
     \begin{table}[H]
         \centering
         \begin{adjustbox}{width=0.6\textwidth}
             \small
             \begin{tabular}{|c|c|c|c|c|}
               \hline    1(Not at all) & 2 & 3 & 4 & 5 (Very) \\ \hline
             \end{tabular}
         \end{adjustbox}
     \end{table}

    \item How Intelligent are you when playing games?
     
     \begin{table}[H]
         \centering
         \begin{adjustbox}{width=0.6\textwidth}
             \small
             \begin{tabular}{|c|c|c|c|c|}
               \hline    1(Not at all) & 2 & 3 & 4 & 5 (Very) \\ \hline
             \end{tabular}
         \end{adjustbox}
     \end{table}

    \item How innovative are you when playing games?
     
     \begin{table}[H]
         \centering
         \begin{adjustbox}{width=0.6\textwidth}
             \small
             \begin{tabular}{|c|c|c|c|c|}
               \hline    1(Not at all) & 2 & 3 & 4 & 5 (Very) \\ \hline
             \end{tabular}
         \end{adjustbox}
     \end{table}

    \item How organized are you when playing games?
     
      \begin{table}[H]
         \centering
         \begin{adjustbox}{width=0.6\textwidth}
             \small
             \begin{tabular}{|c|c|c|c|c|}
               \hline    1(Not at all) & 2 & 3 & 4 & 5 (Very) \\ \hline
             \end{tabular}
         \end{adjustbox}
     \end{table}

    \item How compassionate are you when playing games?
     
      \begin{table}[H]
         \centering
         \begin{adjustbox}{width=0.6\textwidth}
             \small
             \begin{tabular}{|c|c|c|c|c|}
               \hline    1(Not at all) & 2 & 3 & 4 & 5 (Very) \\ \hline
             \end{tabular}
         \end{adjustbox}
     \end{table}

    \item How affectionate are you when playing games?
     
      \begin{table}[H]
         \centering
         \begin{adjustbox}{width=0.6\textwidth}
             \small
             \begin{tabular}{|c|c|c|c|c|}
               \hline    1(Not at all) & 2 & 3 & 4 & 5 (Very) \\ \hline
             \end{tabular}
         \end{adjustbox}
     \end{table}

    \item How sensitive are you when playing games?
     
      \begin{table}[H]
         \centering
         \begin{adjustbox}{width=0.6\textwidth}
             \small
             \begin{tabular}{|c|c|c|c|c|}
               \hline    1(Not at all) & 2 & 3 & 4 & 5 (Very) \\ \hline
             \end{tabular}
         \end{adjustbox}
     \end{table}

    \item How emotional are you when playing games?
     
      \begin{table}[H]
         \centering
         \begin{adjustbox}{width=0.6\textwidth}
             \small
             \begin{tabular}{|c|c|c|c|c|}
               \hline    1(Not at all) & 2 & 3 & 4 & 5 (Very) \\ \hline
             \end{tabular}
         \end{adjustbox}
     \end{table}

    \item How experienced are you with competitive card games?
     
      \begin{table}[H]
         \centering
         \begin{adjustbox}{width=0.6\textwidth}
             \small
             \begin{tabular}{|c|c|c|c|c|}
               \hline    1(Not at all) & 2 & 3 & 4 & 5 (Very) \\ \hline
             \end{tabular}
         \end{adjustbox}
     \end{table}

\end{enumerate}

\subsection{Post-game Questionnaire}

 \begin{enumerate}
    \item Which was the nickname you choose?
 
     \item How competitive were your opponents?
     
      \begin{table}[H]
         \centering
         \begin{adjustbox}{width=0.6\textwidth}
             \small
             \begin{tabular}{|c|c|c|c|c|c|}
               \hline    &1(Not at all) & 2 & 3 & 4 & 5 (Very) \\ \hline
               \hline    Dylan & &  &  &  &  \\ \hline
               \hline    Frankie & &  &  &  &  \\ \hline
               \hline    Evan & &  &  &  &  \\ \hline

             \end{tabular}
         \end{adjustbox}
     \end{table}
 
  \item How ambitious were your opponents?
     
      \begin{table}[H]
         \centering
         \begin{adjustbox}{width=0.6\textwidth}
             \small
             \begin{tabular}{|c|c|c|c|c|c|}
               \hline    &1(Not at all) & 2 & 3 & 4 & 5 (Very) \\ \hline
               \hline    Dylan & &  &  &  &  \\ \hline
               \hline    Frankie & &  &  &  &  \\ \hline
               \hline    Evan & &  &  &  &  \\ \hline

             \end{tabular}
         \end{adjustbox}
     \end{table}

    \item How courageous were your opponents?
     
      \begin{table}[H]
         \centering
         \begin{adjustbox}{width=0.6\textwidth}
             \small
             \begin{tabular}{|c|c|c|c|c|c|}
               \hline    &1(Not at all) & 2 & 3 & 4 & 5 (Very) \\ \hline
               \hline    Dylan & &  &  &  &  \\ \hline
               \hline    Frankie & &  &  &  &  \\ \hline
               \hline    Evan & &  &  &  &  \\ \hline

             \end{tabular}
         \end{adjustbox}
     \end{table}

    \item How decisive were your opponents?
     
      \begin{table}[H]
         \centering
         \begin{adjustbox}{width=0.6\textwidth}
             \small
             \begin{tabular}{|c|c|c|c|c|c|}
               \hline    &1(Not at all) & 2 & 3 & 4 & 5 (Very) \\ \hline
               \hline    Dylan & &  &  &  &  \\ \hline
               \hline    Frankie & &  &  &  &  \\ \hline
               \hline    Evan & &  &  &  &  \\ \hline

             \end{tabular}
         \end{adjustbox}
     \end{table}

    \item How aggressive were your opponents?
     
      \begin{table}[H]
         \centering
         \begin{adjustbox}{width=0.6\textwidth}
             \small
             \begin{tabular}{|c|c|c|c|c|c|}
               \hline    &1(Not at all) & 2 & 3 & 4 & 5 (Very) \\ \hline
               \hline    Dylan & &  &  &  &  \\ \hline
               \hline    Frankie & &  &  &  &  \\ \hline
               \hline    Evan & &  &  &  &  \\ \hline

             \end{tabular}
         \end{adjustbox}
     \end{table}

    \item How creative were your opponents?
     
      \begin{table}[H]
         \centering
         \begin{adjustbox}{width=0.6\textwidth}
             \small
             \begin{tabular}{|c|c|c|c|c|c|}
               \hline    &1(Not at all) & 2 & 3 & 4 & 5 (Very) \\ \hline
               \hline    Dylan & &  &  &  &  \\ \hline
               \hline    Frankie & &  &  &  &  \\ \hline
               \hline    Evan & &  &  &  &  \\ \hline

             \end{tabular}
         \end{adjustbox}
     \end{table}

    \item How intelligent were your opponents?
     
      \begin{table}[H]
         \centering
         \begin{adjustbox}{width=0.6\textwidth}
             \small
             \begin{tabular}{|c|c|c|c|c|c|}
               \hline    &1(Not at all) & 2 & 3 & 4 & 5 (Very) \\ \hline
               \hline    Dylan & &  &  &  &  \\ \hline
               \hline    Frankie & &  &  &  &  \\ \hline
               \hline    Evan & &  &  &  &  \\ \hline

             \end{tabular}
         \end{adjustbox}
     \end{table}

    \item How innovative were your opponents?
     
     \begin{table}[H]
         \centering
         \begin{adjustbox}{width=0.6\textwidth}
             \small
             \begin{tabular}{|c|c|c|c|c|c|}
               \hline    &1(Not at all) & 2 & 3 & 4 & 5 (Very) \\ \hline
               \hline    Dylan & &  &  &  &  \\ \hline
               \hline    Frankie & &  &  &  &  \\ \hline
               \hline    Evan & &  &  &  &  \\ \hline

             \end{tabular}
         \end{adjustbox}
     \end{table}

    \item How organized were your opponents?
     
     \begin{table}[H]
         \centering
         \begin{adjustbox}{width=0.6\textwidth}
             \small
             \begin{tabular}{|c|c|c|c|c|c|}
               \hline    &1(Not at all) & 2 & 3 & 4 & 5 (Very) \\ \hline
               \hline    Dylan & &  &  &  &  \\ \hline
               \hline    Frankie & &  &  &  &  \\ \hline
               \hline    Evan & &  &  &  &  \\ \hline

             \end{tabular}
         \end{adjustbox}
     \end{table}

    \item How compassionate were your opponents?
     
     \begin{table}[H]
         \centering
         \begin{adjustbox}{width=0.6\textwidth}
             \small
             \begin{tabular}{|c|c|c|c|c|c|}
               \hline    &1(Not at all) & 2 & 3 & 4 & 5 (Very) \\ \hline
               \hline    Dylan & &  &  &  &  \\ \hline
               \hline    Frankie & &  &  &  &  \\ \hline
               \hline    Evan & &  &  &  &  \\ \hline

             \end{tabular}
         \end{adjustbox}
     \end{table}

    \item How affectionate were your opponents?
     
     \begin{table}[H]
         \centering
         \begin{adjustbox}{width=0.6\textwidth}
             \small
             \begin{tabular}{|c|c|c|c|c|c|}
               \hline    &1(Not at all) & 2 & 3 & 4 & 5 (Very) \\ \hline
               \hline    Dylan & &  &  &  &  \\ \hline
               \hline    Frankie & &  &  &  &  \\ \hline
               \hline    Evan & &  &  &  &  \\ \hline

             \end{tabular}
         \end{adjustbox}
     \end{table}

    \item How sensitive  were your opponents?
     
     \begin{table}[H]
         \centering
         \begin{adjustbox}{width=0.6\textwidth}
             \small
             \begin{tabular}{|c|c|c|c|c|c|}
               \hline    &1(Not at all) & 2 & 3 & 4 & 5 (Very) \\ \hline
               \hline    Dylan & &  &  &  &  \\ \hline
               \hline    Frankie & &  &  &  &  \\ \hline
               \hline    Evan & &  &  &  &  \\ \hline

             \end{tabular}
         \end{adjustbox}
     \end{table}

    \item How emotional were your opponents?
     
     \begin{table}[H]
         \centering
         \begin{adjustbox}{width=0.6\textwidth}
             \small
             \begin{tabular}{|c|c|c|c|c|c|}
               \hline    &1(Not at all) & 2 & 3 & 4 & 5 (Very) \\ \hline
               \hline    Dylan & &  &  &  &  \\ \hline
               \hline    Frankie & &  &  &  &  \\ \hline
               \hline    Evan & &  &  &  &  \\ \hline

             \end{tabular}
         \end{adjustbox}
     \end{table}
 
     \item Would you like to play with these opponents again?
     
     \begin{table}[H]
         \centering
         \begin{adjustbox}{width=0.3\textwidth}
             \small
             \begin{tabular}{|c|c|c|}
               \hline    &Yes & No  \\ \hline
               \hline    Dylan & & \\ \hline
               \hline    Frankie & &    \\ \hline
               \hline    Evan & &   \\ \hline

             \end{tabular}
         \end{adjustbox}
     \end{table}
 
 \end{enumerate}
 
 \section{Similarity Predictor Optimization}
 
 To optimize this network we used a Tree-Parzen Optimizer (Hyperopt \citep{bergstra2013hyperopt}) based on the search space defined in Table \ref{tab:searchSpaceSimilarity}.
 
\begin{table}[H]
\center
\begin{tabular}{ |c|c|} 
\hline
 \textbf{Parameter} & \textbf{Search Space} \\ \hline
 \multicolumn{2}{|c|}{\textbf{Similarity Predictor}}\\\hline
 Number of Concatenated Actions & [3,5,10, 15]\\
 Number of Layers & [1,2,3]\\
 Units Per Layer & [16, 32, 64, 128, 256, 512, 1024] \\
 \hline
\end{tabular}
\caption{Search space used to optimize the similarity predictor neural network. }
\label{tab:searchSpaceSimilarity}
\end{table}




\end{appendices}


\bibliography{sn-bibliography}


\end{document}